\documentclass{article}

\usepackage{arxiv}

\usepackage[utf8]{inputenc}
\usepackage[T1]{fontenc}
\usepackage{hyperref}
\usepackage{url}
\usepackage{booktabs}
\usepackage{amsfonts}
\usepackage{amsmath}
\usepackage{graphicx}
\usepackage{subcaption}
\usepackage[table]{xcolor}
\usepackage[numbers]{natbib}
\usepackage{doi}

\title{Understanding How MLLMs Describe Artworks Using Token Activation Maps}

\renewcommand{\shorttitle}{Understanding How MLLMs Describe Artworks}

\newcommand{\authorblock}[1]{\begin{minipage}[t]{0.31\textwidth}\centering #1\end{minipage}}
\author{
  \authorblock{
  \href{https://orcid.org/0009-0007-6602-7504}{Nicola Fanelli}\thanks{Funded by a Ph.D. fellowship under the Italian ``D.M. n. 118/23'' (NRRP, Mission 4, Investment 4.1, CUP H91I23000690007).} \\
  Department of Computer Science\\
  University of Bari Aldo Moro\\
  Bari, Italy\\
  \texttt{nicola.fanelli@uniba.it}}
  \And
  \authorblock{
  \href{https://orcid.org/0000-0001-8935-9156}{Pasquale De Marinis} \\
  Department of Computer Science\\
  University of Bari Aldo Moro\\
  Bari, Italy\\
  \texttt{pasquale.demarinis@uniba.it}}
  \And
  \authorblock{
  \href{https://orcid.org/0000-0001-7512-7661}{Raffaele Scaringi} \\
  Department of Computer Science\\
  University of Bari Aldo Moro\\
  Bari, Italy\\
  \texttt{raffaele.scaringi@uniba.it}}
  \AND
  \authorblock{
  \href{https://orcid.org/0000-0002-5330-1259}{Eva Cetinic} \\
  Digital Society Initiative\\
  University of Zurich\\
  Zurich, Switzerland\\
  \texttt{eva.cetinic@uzh.ch}}
  \And
  \authorblock{
  \href{https://orcid.org/0000-0002-0883-2691}{Gennaro Vessio} \\
  Department of Computer Science\\
  University of Bari Aldo Moro\\
  Bari, Italy\\
  \texttt{gennaro.vessio@uniba.it}}
  \And
  \authorblock{
  \href{https://orcid.org/0000-0002-6489-8628}{Giovanna Castellano} \\
  Department of Computer Science\\
  University of Bari Aldo Moro\\
  Bari, Italy\\
  \texttt{giovanna.castellano@uniba.it}}
}

\hypersetup{
pdftitle={Understanding How MLLMs Describe Artworks Using Token Activation Maps},
pdfauthor={Nicola Fanelli, Pasquale De Marinis, Raffaele Scaringi, Eva Cetinic, Gennaro Vessio, Giovanna Castellano},
pdfkeywords={Multimodal Large Language Models, Explainable AI, Art understanding},
}

\begin{document}
% Zero the tabular padding only around \maketitle so three fixed-width author
% boxes fit on one line; restored immediately after for the body tables.
{\setlength{\tabcolsep}{0pt}\maketitle}

\begin{abstract}
Multimodal Large Language Models (MLLMs) describe artworks with remarkable fluency, yet the visual reasoning behind their outputs remains opaque. When an MLLM names a style, identifies a subject, or recognizes an iconographic symbol, does it ground each claim in the relevant region of the canvas, draw on an undifferentiated visual signal, or rely primarily on textual priors? We study this using the Token Activation Map (TAM), which produces, for each generated token, a heatmap isolating the visual evidence specific to that token from prior-context interference. Applying TAM to a curated set of paintings spanning multiple periods and genres, we analyze grounding patterns across five semantically distinct token categories: common visual objects, style descriptors, metadata, iconographic tokens, and affective expressions. We find that visual grounding varies substantially with token semantics. We further show that MLLMs attempt to identify artworks and artists, achieving higher accuracy in artist attribution than in title prediction, where hallucinations are more frequent. Finally, we compare TAM with SAM~3 open-vocabulary segmentation. To ensure reproducibility, we release our code, experimental configurations,
prompts, and qualitative results on the
project page at \href{https://nicolafan.github.io/tamart/}{\textcolor{blue}{\texttt{https://nicolafan.github.io/tamart/}}}.
% , report failure cases, and outline directions for future interpretability-driven evaluation in cultural heritage applications.
\end{abstract}


\begin{figure}[t]
    \centering
    \includegraphics[width=1.0\linewidth]{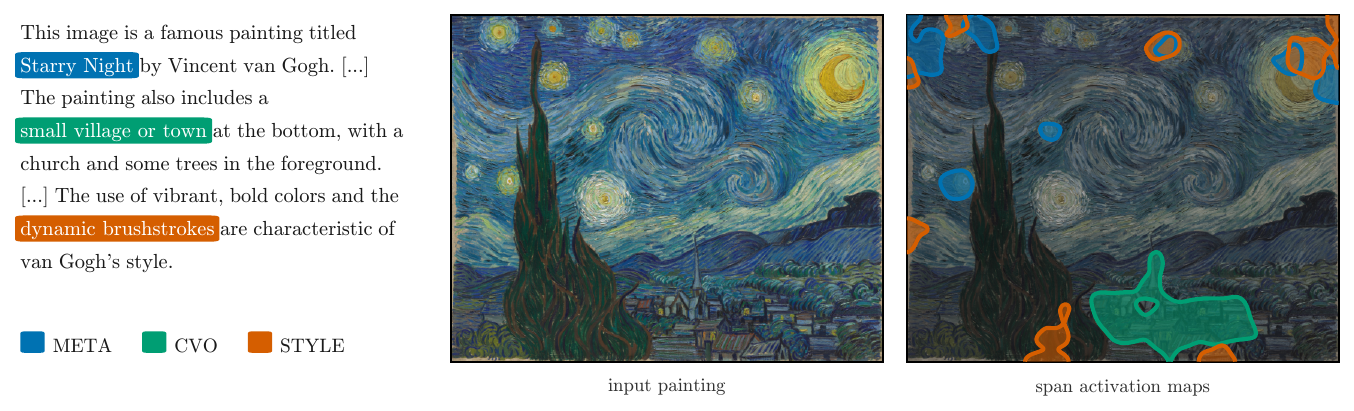}
    \caption{\textbf{We present a token-level view of how multimodal LLMs \emph{see} the art they describe.} Using Token Activation Maps, we trace each generated span back to the image region it draws on. Grounding depends on \emph{what} is said: a concrete subject (\textsc{cvo}, \emph{``small village or town''}) localizes to one region, while a style descriptor (\textsc{style}, \emph{``dynamic brushstrokes''}) and a metadata mention (\textsc{meta}, \emph{``Starry Night''}) spread diffusely across the canvas.}
    \label{fig:intro}
\end{figure}

\section{Introduction}

Multimodal Large Language Models (MLLMs)~\cite{wu2023multimodal} have significantly advanced the state of the art in image understanding. These models can describe visual scenes, answer open-ended questions, and perform multimodal reasoning with high degree of fluency. Architectures such as LLaVA~\cite{liu2023visual}, Qwen3-VL~\cite{bai2025qwen3}, and InternVL~\cite{chen2024internvl} integrate visual encoders with autoregressive language models, allowing them to generate detailed natural-language interpretations of images. Beyond generic visual tasks, these models are increasingly  applied and explored in contexts related to cultural heritage, where they demonstrate promising capabilities in artwork captioning, stylistic analysis, and visual question answering.

% In the context of computational image understanding, artworks are of particular interest because they introduce challenges that differ substantially from those posed by photorealistic imagery. Understanding a painting requires not only the recognition of concrete visual entities, but also the interpretation of painterly style, symbols, historical context, and affective atmosphere~\cite{cetinic2019deep}. A single artwork caption may simultaneously refer to localized objects (e.g., “a woman holding a child”), global stylistic properties (e.g., “loose impressionistic brushwork”), emotional interpretations (e.g., “a melancholic atmosphere”), and metadata (e.g., “painted by Vincent van Gogh”). As a result, understanding artwork requires both low-level visual perception and high-level semantic interpretation.

% Despite the fluency of modern MLLMs, the visual grounding underlying their outputs remains insufficiently understood. When a model attributes a painting to a specific artist, identifies an iconographic figure, or describes a stylistic characteristic, it is often unclear whether the generated claim is grounded in relevant image regions or instead reflects language priors acquired during training. This limitation is particularly critical in cultural heritage applications, where interpretability and trustworthiness are essential.

Artworks pose challenges that differ substantially from photorealistic imagery:
understanding a painting requires not only recognition of concrete visual entities,
but also interpretation of style, iconographic symbols, and affective
atmosphere~\cite{cetinic2019deep}. A single caption may simultaneously refer to
localized objects (e.g., ``a woman holding a child''), global stylistic properties
(e.g., ``loose impressionistic brushwork''), emotional tone (e.g., ``a melancholic
atmosphere''), and authorship metadata (e.g., ``painted by Vincent van Gogh'').
Despite the fluency of modern MLLMs, the visual grounding underlying their outputs
remains poorly understood---it is often unclear whether a generated claim is
anchored in relevant image regions or merely reflects language priors acquired
during training. This limitation is particularly critical in cultural heritage
applications, where interpretability and trustworthiness are essential.

To address this issue, a large body of work has investigated explainability in deep vision models through methods such as CAM~\cite{zhou2016learning}, Grad-CAM~\cite{selvaraju2017grad}, attention rollout~\cite{abnar2020quantifying}, and concept-based attribution techniques such as TCAV~\cite{kim2018interpretability}. Within the art domain, several studies have applied explainability methods to analyze visual models for tasks such as emotion recognition and multi-task artwork classification~\cite{castellano2024using,scaringi2025graphclip}. However, these approaches typically focus on classification outputs and provide a single global explanation per image. Consequently, they do not capture the token-by-token nature of reasoning in autoregressive multimodal generation.

To overcome this limitation, the Token Activation Map (TAM) approach~\cite{li2025token} was introduced to isolate token-specific visual evidence while mitigating interference from prior generation context. In this work, we leverage TAM to investigate how MLLMs visually ground different semantic components of artwork descriptions. We apply this framework to captions generated for paintings spanning multiple artistic periods and genres, and analyze the spatial characteristics of activation maps across five semantically distinct categories (Fig.~\ref{fig:intro}). Our goal is to understand whether different aspects of artwork interpretation correspond to distinct patterns of visual grounding, and whether abstract concepts such as style and affect rely more on dispersed visual cues and patterns than concrete depicted entities.

% Our analysis reveals substantial differences in grounding behavior across semantic categories. We show that activation maps associated with concrete objects and iconographic figures are significantly more localized than those corresponding to stylistic or affective descriptions, which tend to activate more broadly across the image. We further examine metadata-related spans and find that MLLMs are considerably more reliable at artist attribution than artwork title prediction, where hallucinations are more frequent. Finally, we compare TAM-based explanations with open-vocabulary segmentation methods and discuss implications for explainable AI in cultural heritage applications.

The rest of this paper is structured as follows. Section~\ref{sec:rel-works} discusses related work. Section~\ref{sec:methodology} describes the proposed methodology. Section~\ref{sec:experiments} presents experimental results and evaluation. Section~\ref{sec:conclusions} concludes the paper and outlines directions for future research.

\section{Related Work}\label{sec:rel-works}

\subsection{Vision-Language Models for Art Understanding}

Interest in applying vision-language models to art understanding predates the MLLM era. Early work on semantic art retrieval~\cite{garcia2018read} and multimodal visual question answering about artworks~\cite{garcia2020dataset} established the core benchmarks for the field. Contrastive VLMs such as CLIP were subsequently probed on art-domain tasks~\cite{scaringi2025graphclip}, showing effective stylistic clustering~\cite{conde2021clip} and sensitivity to high-level formal categories such as W\"olfflin's principles~\cite{ghildyal2025wp}, while also revealing limited perception of period and aesthetic nuance~\cite{asperti2025does}.

The emergence of MLLMs has significantly expanded the possibilities for art understanding. Generalist systems such as LLaVA~\cite{liu2023visual}, Qwen3-VL~\cite{bai2025qwen3}, and InternVL~\cite{chen2024internvl} provide strong baselines, and several works have adapted or evaluated them specifically in the art domain. GalleryGPT~\cite{bin2024gallerygpt} fine-tuned LLaVA on formal analysis annotations and exposed systematic failure modes in GPT-4V on canonical paintings. VQArt-Bench~\cite{alfarano2025vqart} constructed an evaluation suite targeting iconographic and symbolic reasoning, revealing significant gaps between proprietary and open-source MLLMs. On the generative side, I Dream My Painting~\cite{fanelli2025dream} connects MLLMs with diffusion-based inpainting for artwork restoration, NADA~\cite{ramos2025no} harnesses cross-attention maps for annotation-free iconographic detection, and ArtSeek~\cite{fanelli2025artseek} integrates late-interaction retrieval with Qwen2.5-VL for grounded, knowledge-augmented artwork description.

Despite this growing body of work, none of these approaches examines \emph{which visual regions} an MLLM attends to when generating different aspects of an artwork description, which is the central question addressed in this paper.

\subsection{Explainability of Vision and Multimodal Models}

Interpretability of deep vision models has evolved from gradient-weighted class
activation maps~\cite{selvaraju2017grad} and concept-based probing~\cite{kim2018interpretability}
to attention rollout~\cite{abnar2020quantifying} and relevance-propagation methods
for ViTs~\cite{chefer2021transformer}. For contrastive vision-language models,
multi-modal attribution~\cite{wang2023visual} produces joint image-text relevance
maps, Visual-TCAV~\cite{achtibat2023attribution} extends integrated-gradient
attribution to per-concept localization, and SpLiCE~\cite{bhalla2024interpreting}
decomposes CLIP embeddings into sparse semantic concepts without task-specific
training.
Interpreting autoregressive MLLMs is harder, since each output token is
generated conditioned on all prior context, making naive activation maps noisy
and confounded. The logit-lens family partially addresses this by projecting
intermediate activations into the vocabulary space~\cite{belrose2023eliciting},
and subsequent probing work finds that visual-to-class projection concentrates
in specific mid-to-late LLM layers~\cite{neo2025towards}. TAM~\cite{li2025token}
resolves the confounding issue more directly: for each generated token it estimates
an interference map from earlier context via least-squares subtraction, then
applies a rank Gaussian filter to suppress residual noise, yielding a per-token
heatmap that localizes the image regions most relevant to each generated word.
Evaluated across seven MLLM architectures, TAM establishes state-of-the-art
localization performance and provides the methodological basis for the present work.

\subsection{Explainability Applied to the Analysis of Artworks}

Despite the wide range of existing interpretability methods, their application to the art domain remains sparse. In the context of CNN-based classification, foundational studies using Grad-CAM and its variants~\cite{pinciroli2021comparing} directly compared CAM, Grad-CAM, Grad-CAM++, and Smooth Grad-CAM++ on a neural network trained on Christian iconography, showing that the maps highlight iconographically meaningful regions---attributes such as halos, lambs, and arrows---with bounding boxes reaching 55\% mean Intersection-over-Union (IoU) against manually annotated iconographic elements. The ArtDL dataset and accompanying ResNet baseline~\cite{milani2021dataset} similarly confirmed through CAM visualizations that the model focuses on expected iconographic attributes for individual saint classes. Shifting to transformer architectures, Diem and Mandl~\cite{diem2023automatic} contrasted Grad-CAM on a ResNet with attention maps on a ViT for artist-attribution of portrait prints, finding that the ViT attends almost exclusively to the depicted person while the CNN concentrates on local printing-technique textures. Strafforello et al.~\cite{strafforello2025have} conducted the first evaluation of generative VLMs on art style, author, and period classification in a zero-shot regime. Most directly related to the present work, Schneider~\cite{schneider2026explainability} benchmarks seven XAI methods on CLIP in art-historical contexts and finds that the legibility of visual reasoning depends heavily on the conceptual stability and representational availability of the examined categories, while Limpijankit et al.~\cite{limpijankit2026does} decompose the latent space of VLMs to identify the concepts driving art style prediction, reporting that 73\% of extracted concepts were judged coherent by art historians. Both works, however, are anchored in contrastive VLMs and operate at the level of a single classification output rather than token-by-token generation of a full textual description.

Our work takes a complementary step: by applying TAM to a generative MLLM, we obtain a distinct attribution map for \emph{each} word in the painting description, enabling a systematic comparison of the visual grounding of style-related, content-related, and iconographic tokens, and asking whether an MLLM partitions its visual attention across these semantically distinct facets of art in a way that aligns with art-historical reasoning.
% \nicola{Find a way to explain he difference between VLMs and MLLMs. Literature is not super clear about this but VLM -> any vision-lagnuage model, such as CLIP or (almost) every MLLM, MLLM -> (not only but especially) vision-langauge models for text generation that have an LLM inside. I don't know where to put this}

\section{Methodology}
\label{sec:methodology}

We describe a pipeline that, given a digitized artwork, produces a set of semantically typed visual activation maps revealing \emph{where} in the image an MLLM grounds different aspects of its generated description---concrete objects, iconographic subjects, stylistic attributes, affective judgments, and metadata. The pipeline proceeds in four stages: we first describe the data, model, and caption generation procedure (Section~\ref{sec:method:data-model-captioning}); we then summarize the Token Activation Map (TAM) method we adopt for token-level visual grounding (Section~\ref{sec:method:tam}); we classify the generated captions into semantically typed spans (Section~\ref{sec:method:spans}); and finally we aggregate the token-level maps into span-level maps that serve as the basic unit of analysis (Section~\ref{sec:method:aggregation}).

\subsection{Data, Model, and Caption Generation}
\label{sec:method:data-model-captioning}

To study how an MLLM grounds its description of a painting, we  assembled a corpus by collecting the 1{,}000 most-viewed paintings from WikiArt\footnote{\url{https://www.wikiart.org}} through its public API, downloaded at the platform's maximum resolution. Popularity is a useful proxy here: the most-viewed works skew toward famous paintings, enabling us to probe---through the \textsc{meta} category introduced in Section~\ref{sec:method:spans}---whether the model recovers correct metadata for recognizable works, and whether its visual grounding shifts between familiar and unfamiliar ones.

We paired this corpus with Qwen2-VL-2B-Instruct~\cite{wang2024qwen2} as our MLLM backbone. An MLLM couples a Vision Transformer~\cite{dosovitskiy2021an} encoder to an autoregressive LLM through a lightweight projector: the encoder turns the image into patch-level visual tokens, the projector maps them into the LLM's embedding space, and they are concatenated with the tokenized prompt before generation proceeds token by token. What matters for our purposes is that, at every generation step, the hidden state at the last layer aggregates information from the image, the prompt, and all previously generated tokens---the property that the method of Section~\ref{sec:method:tam} will exploit to recover \emph{where} in the image each generated token is grounded. Qwen2-VL is well-suited to this setting because it preserves fine-grained spatial
information more carefully than most MLLMs: it processes images at their native
resolution via \emph{Naive Dynamic Resolution} rather than rescaling to a fixed
size, and it injects spatial structure directly into attention through 2D Rotary
Position Embeddings~\cite{heo2024rotary}. In practice, each image is rescaled
before encoding to fall within a range of $256$ to $1{,}280$ visual tokens, at
one token per $28\!\times\!28$ pixel patch, so that the patch
grid---to which our activation maps will ultimately be aligned---remains faithful
to the original image. Its 2B parameter scale also offers a practical balance
between representational capacity and computational cost. Furthermore,
Qwen2-VL-2B-Instruct is the primary model on which TAM is developed and
benchmarked~\cite{li2025token}, and Li et al.\ show that the method yields
consistent improvements across architecturally distinct MLLMs---including LLaVA-1.5
and InternVL2.5~\cite{li2025token}---with no evidence of qualitatively different
grounding behaviour between families; it thus serves as a reasonable proxy for
the broader class of decoder-based multimodal models, with extension to further
architectures left as future work.

With the corpus and the model in place, generating descriptions is straightforward. For each painting $\mathbf{I}$, we prompt the model with \texttt{``Describe the content and style of this image.''} and let it autoregressively produce an answer sequence $t^a_1, \dots, t^a_{n_a}$ of up to $256$ tokens. The prompt is intentionally open-ended: by asking for both \emph{content} and \emph{style}, it invites the model to range over concrete visual objects, stylistic observations, affective impressions, and---for recognized paintings---iconographic references and metadata such as artist or title. As the model generates, we extract a per-token visual activation map for every $t^a_i$ using the procedure described next.

\subsection{Token-Level Visual Grounding with TAM}
\label{sec:method:tam}

To localize the contribution of each generated token to the image, we adopt the Token Activation Map method of Li et al.~\cite{li2025token}. Classical activation-mapping techniques such as CAM~\cite{zhou2016learning} and Grad-CAM~\cite{selvaraju2017grad} target classifiers that emit a single prediction, and applying them token-by-token to an autoregressive MLLM yields noisy and unreliable maps. TAM is designed precisely for this setting, and we summarize here only the intuitions needed to follow Section~\ref{sec:method:aggregation}; we refer the reader to the original paper for the full formulation.

The starting point is a \emph{raw} activation map for each generated token $t^a_i$, obtained by projecting the visual patch features against the row of the LLM's output classifier corresponding to that token. Intuitively, this measures how strongly each image patch aligns with the concept the token encodes---the same principle as CAM, applied independently at every generation step. The difficulty is that these raw maps are systematically corrupted. Because generation is autoregressive, the hidden state that produces $t^a_i$ has already absorbed information from the image, the prompt, and every earlier generated token, so the raw map for $t^a_i$ inherits \emph{interference} from correlated context tokens that activate similar image regions. If a caption first mentions a \emph{horse} and then names its \emph{rider}, the raw map for ``rider'' tends to bleed onto the body of the horse, since the two entities are spatially adjacent and the hidden state for ``rider'' still carries information about the earlier token.

TAM removes this interference by treating each earlier token's raw map as a potential confounder and subtracting an estimated interference map---a relevance-weighted combination of the context maps.
The weights are computed from textual similarity between the context tokens and $t^a_i$; identical tokens are excluded so the model does not suppress genuine self-activation. The optimal subtraction scale is fit in closed form by least squares. A rank Gaussian filter is then applied to suppress residual salt-and-pepper noise,
yielding for each generated token $t^a_i$ a refined visual activation map
$\bar{\mathbf{A}}^{a}_{i} \in [0,1]^{n_v}$ defined over the $n_v$
patches of the visual encoder's grid. This is the only TAM output we use in what
follows: by construction, $\bar{\mathbf{A}}^{a}_{i}$ encodes \emph{where} the
model grounds token $t^a_i$ in the image, with the contribution of correlated
context tokens already removed.

\subsection{Semantic Span Classification}
\label{sec:method:spans}

An MLLM description of a painting interleaves tokens of very different semantic nature: a single sentence may name a depicted figure, comment on a color palette, ascribe a mood, and recall the artist's name. To disentangle these, we classify contiguous token spans of the generated caption into five categories:
\begin{itemize}
    \item \textbf{\textsc{cvo}} (Concrete Visual Object): a depicted entity or its visible attribute that can be localized in the image, e.g.\ \emph{``woman''}, \emph{``body of water''};
    \item \textbf{\textsc{icon}} (Iconographic Subject): a named mythological, religious, or historical figure depicted in the scene, e.g., \emph{``Dante''}, \emph{``Madonna''};
    \item \textbf{\textsc{style}} (Painterly Attribute): a reference to brushwork, palette, technique, or period/movement, e.g., \emph{``loose brushwork''}, \emph{``Renaissance''};
    \item \textbf{\textsc{affect}} (Affective/Interpretive): a mood, emotion, or atmospheric quality, e.g., \emph{``dramatic atmosphere''}, \emph{``serene''};
    \item \textbf{\textsc{meta}} (Metadata): artist name, painting title, date, or provenance, e.g., \emph{``Leonardo da Vinci''}, \emph{``1642''}.
\end{itemize}

Classification is performed by a separate text-only instruct LLM, Qwen3-4B-Instruct-2507~\cite{bai2025qwen3}, which we also reuse as an LLM-as-a-judge in Section~\ref{sec:exp2}. The caption is presented to the classifier as an indexed token list---each token
prefixed by its position in the generated sequence---and the model is prompted to
return a structured JSON response specifying, for each detected expression, its
surface form, the span of token indices $\{j, \dots, j{+}l{-}1\}$ it occupies
(where $l$ denotes the expression's length in tokens), and its assigned category. The indexed presentation lets the model report token positions directly, sidestepping the need to align surface forms back to the tokenized sequence post hoc. We cap each expression at ten tokens and discard purely positional fillers (e.g., \emph{``in the foreground''}) to focus on contentful spans. This yields a set of typed spans $\mathcal{S} = \{(e_m, \mathcal{I}_m, y_m)\}_{m=1}^{M}$, where $e_m$ is the expression text, $\mathcal{I}_m \subseteq [1, n_a]$ is the set of token indices, and $y_m \in \{\textsc{cvo}, \textsc{icon}, \textsc{style}, \textsc{affect}, \textsc{meta}\}$ is the category. Applied to the full corpus, this procedure produces $12{,}878$ labeled spans, which---together with the per-token activation maps---are precomputed once and reused across all experiments.

\subsection{Span-Level Map Aggregation}
\label{sec:method:aggregation}

TAM produces one activation map per token, but many of the semantic concepts identified in Section~\ref{sec:method:spans} span multiple tokens---\emph{loose brushwork} and \emph{Leonardo da Vinci}, for instance, are each tokenized into several sub-word units. Li et al.~\cite{li2025token} address this for evaluation purposes by selecting, among the tokens of a multi-token word, the one whose map achieves the highest IoU with a ground-truth segmentation mask. In our setting, no such ground truth is available, so an IoU-based selection criterion is inapplicable.

We instead define the span map as the \emph{element-wise average} of the refined per-token maps over the span's constituent tokens:
\begin{equation}
    \bar{\mathbf{A}}^{\mathrm{span}}_{m}
    = \frac{1}{|\mathcal{I}_m|}
      \sum_{i \in \mathcal{I}_m} \bar{\mathbf{A}}^{a}_{i}.
    \label{eq:span_map}
\end{equation}
We compared this averaging strategy against two natural alternatives---taking only the first token's map, and taking the per-element maximum across the span---through an informal qualitative inspection on a subset of our captions. Averaging produced the most spatially coherent maps in our setting: first-token selection can under-represent the span's grounding when the semantically distinctive sub-word appears later (a frequent pattern in BPE-style tokenization), while the per-element maximum tends to over-extend activations by inheriting spurious peaks from individual tokens.

The resulting span map $\bar{\mathbf{A}}^{\mathrm{span}}_{m}$, paired with its category label $y_m$, is the basic unit of analysis in the experiments of Section~\ref{sec:experiments}: it lets us inspect, for instance, whether \textsc{style} spans activate diffusely across the canvas while \textsc{cvo} spans concentrate on localized regions, or whether \textsc{affect} tokens are grounded in specific visual elements rather than responding to the image holistically.

% Intro to the Experiments section. Citations needed: WikiArt, Qwen2-VL, % Qwen3, vLLM. \ref labels: sec:method (where the span categories are defined), % sec:exp2 (the metadata experiment).

\section{Experiments}
\label{sec:experiments}

We evaluated the proposed pipeline by analyzing semantic variation in visual grounding, metadata reliability, and relation to another localization method.

\subsection{Spatial Localization of Activation Maps by Content Type}
\label{sec:exp1}

We investigated whether Token Activation Maps localize differently depending on the \emph{type} of content a token describes. For every span, we averaged the TAMs of its tokens and measured the \emph{normalized spatial entropy} of the resulting map---the Shannon entropy of the map, viewed as a distribution over the $N$ vision-grid cells, divided by $\log N$---which lies in $[0,1]$, with $1$ fully diffuse and $0$ fully localized. We complemented entropy with two concentration measures: the Gini coefficient of the map values and the fraction of activation in the hottest $10\%$ of cells.

\begin{figure}[t]
  \centering
  \begin{subfigure}[b]{0.32\linewidth}
    \includegraphics[width=\linewidth]{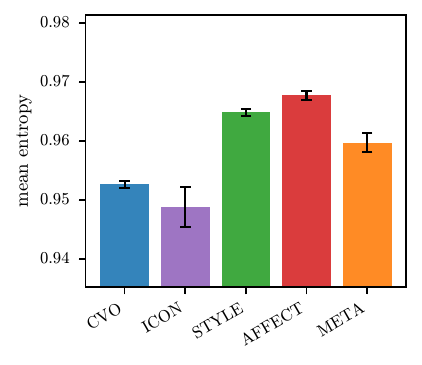}
    \caption{}\label{fig:exp1a}
  \end{subfigure}\hfill
  \begin{subfigure}[b]{0.32\linewidth}
    \includegraphics[width=\linewidth]{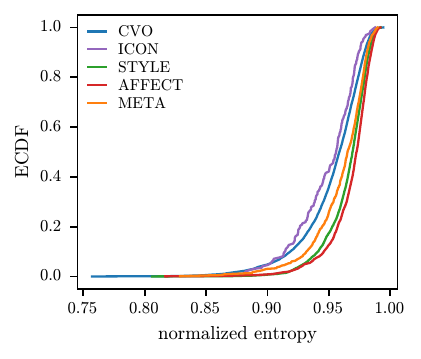}
    \caption{}\label{fig:exp1b}
  \end{subfigure}\hfill
  \begin{subfigure}[b]{0.32\linewidth}
    \includegraphics[width=\linewidth]{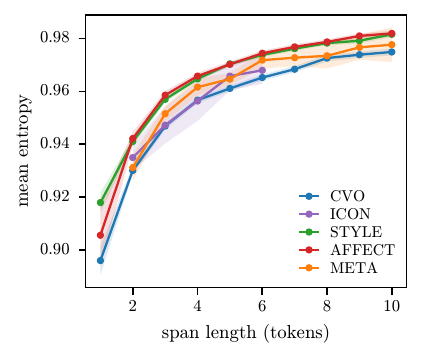}
    \caption{}\label{fig:exp1c}
  \end{subfigure}
  \caption{Activation maps localize by content type. (a) Mean normalized spatial entropy per span type ($95\%$ CI); lower is more localized. (b) Cumulative distributions of entropy. (c) Mean entropy vs.\ exact span length: the ordering holds at every length, so it is not an artifact of token-averaging.}
  \label{fig:exp1}
\end{figure}

Concrete content is significantly more localized than abstract content (Table~\ref{tab:exp1}). Mean entropy is lowest for \textsc{cvo} ($0.953$) and \textsc{icon} ($0.949$) and highest for \textsc{style} ($0.965$) and \textsc{affect} ($0.968$), with \textsc{meta} ($0.960$) in between; the Gini coefficient and top-$10\%$ mass ranked the categories in exactly the opposite order, confirming the metric. Figure~\ref{fig:exp1a} reports the per-type means with $95\%$ confidence intervals and Figure~\ref{fig:exp1b} the cumulative distributions, which are cleanly stochastically ordered. A Kruskal--Wallis test rejected equality of the distributions ($p < 0.001$), and the effect survived controlling for within-painting dependence (paired Friedman test over the paintings containing every category, $p < 0.001$).

\begin{figure}[!t]
  \centering
  \includegraphics[width=\linewidth]{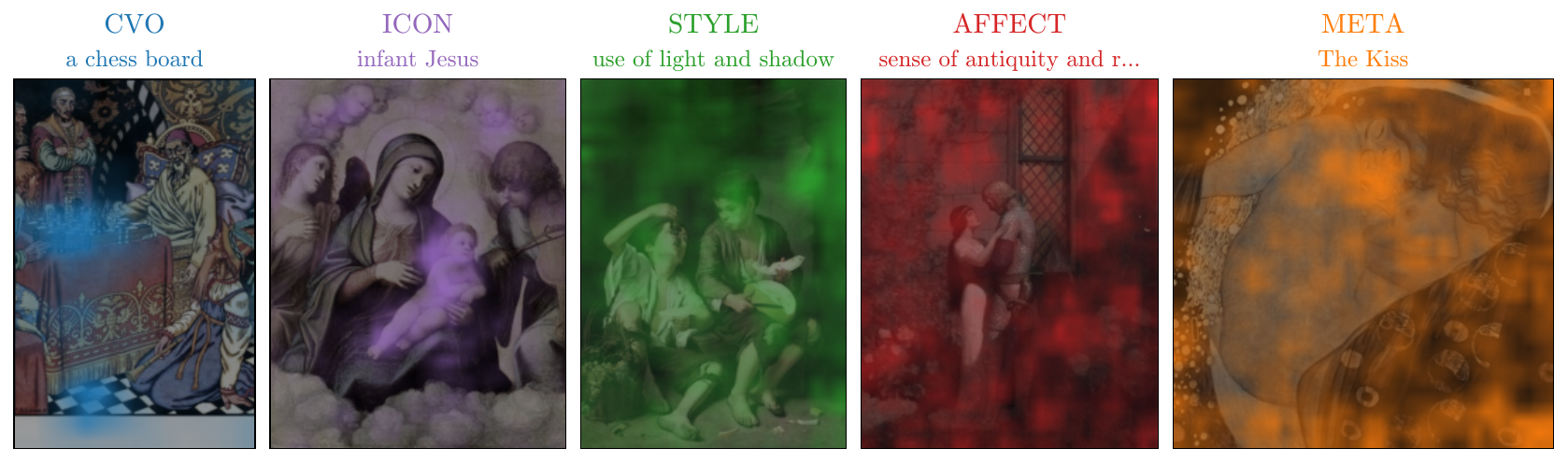}
  \caption{Representative token activation maps, one per content type. Concrete objects (\textsc{cvo}) and named icons (\textsc{icon}) localize tightly, while style (\textsc{style}) and affect (\textsc{affect}) spans are diffuse; \textsc{meta} lies in between. Notice how the \textsc{meta} span \emph{``The Kiss''} corresponds to a model hallucination.}
  \label{fig:exp1examples}
\end{figure}

Because a span map is the mean of its token maps, longer spans average more maps and are mechanically biased toward higher entropy, and span length correlates with both category and entropy (Spearman $\rho \approx 0.72$). We therefore confirmed that the effect is not a length artifact. Figure~\ref{fig:exp1c} plots mean entropy against the \emph{exact} span length: the per-type curves stayed separated and ordered at every length, including single-token spans, for which no averaging occurs. With length held fixed, the categories still differed ($p < 0.001$ within each length), and regressing out the length trend left the ordering intact (Kruskal--Wallis on residuals, $p < 0.001$). Figure~\ref{fig:exp1examples} shows one representative map per type: \textsc{cvo} and \textsc{icon} spans concentrate on a single object, whereas \textsc{style} and \textsc{affect} spans draw on the whole canvas.

% needs \usepackage[table]{xcolor} and \usepackage{booktabs}
\definecolor{loccol}{HTML}{2B7BBA}
\begin{table}[t]
\centering
\caption{Localization of token activation maps by content type ($N=12{,}878$ spans). Normalized spatial entropy, the Gini coefficient, and top-$10\%$ mass concentration. Arrows indicate the direction of stronger localization; cells are shaded so that darker indicates greater localization. Concrete content (\textsc{cvo}, \textsc{icon}) is the most localized, abstract content (\textsc{style}, \textsc{affect}) is the most diffuse.}
\label{tab:exp1}
\begin{tabular*}{\linewidth}{@{\extracolsep{\fill}}lrccc@{}}
\toprule
Span type & $n$ & Entropy~$\downarrow$ & Gini~$\uparrow$ & Top-10\%~$\uparrow$ \\
\midrule
\textsc{cvo} & 6241 & \cellcolor{loccol!46} $0.953\,{\pm}\,0.026$ & \cellcolor{loccol!44} 0.421 & \cellcolor{loccol!47} 0.303 \\
\textsc{icon} & 197 & \cellcolor{loccol!55} $0.949\,{\pm}\,0.024$ & \cellcolor{loccol!55} 0.443 & \cellcolor{loccol!55} 0.312 \\
\textsc{style} & 3542 & \cellcolor{loccol!15} $0.965\,{\pm}\,0.018$ & \cellcolor{loccol!17} 0.368 & \cellcolor{loccol!12} 0.264 \\
\textsc{affect} & 2172 & \cellcolor{loccol!8} $0.968\,{\pm}\,0.018$ & \cellcolor{loccol!8} 0.350 & \cellcolor{loccol!8} 0.259 \\
\textsc{meta} & 726 & \cellcolor{loccol!28} $0.960\,{\pm}\,0.023$ & \cellcolor{loccol!29} 0.392 & \cellcolor{loccol!28} 0.282 \\
\bottomrule
\end{tabular*}
\end{table}

\begin{figure}[t]
  \centering
  \includegraphics[width=\linewidth]{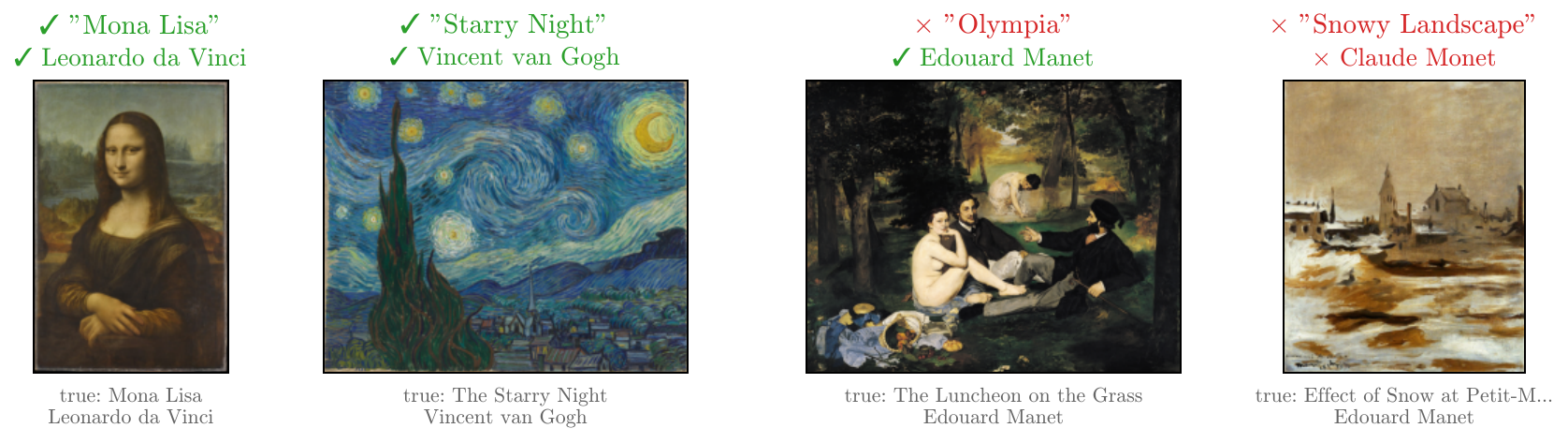}
  \caption{Title and artist predictions extracted from \textsc{meta} spans and graded by an LLM judge (green: correct, red: wrong), with the ground-truth metadata below each painting. Left to right: both correct (\emph{Mona Lisa}, \emph{The Starry Night}); artist correct but title wrong (Manet's \emph{Luncheon on the Grass}, guessed as \emph{Olympia}, another Manet); and both wrong (a Manet snow scene attributed to Monet).}
  \label{fig:exp2examples}
\end{figure}

\subsection{Title and Artist Prediction in \textsc{meta} Spans}
\label{sec:exp2}

A recurring behavior was that some \textsc{meta} spans went beyond providing general background about the artwork and explicitly stated the \emph{title} or the \emph{artist}, which, although confidently phrased, often appeared to be guessed from contextual cues. To assess how reliable these descriptions were, we used an LLM-as-a-judge: for every artwork, the judge read its \textsc{meta} spans, extracted the predicted title and artist when present, and compared each against the ground-truth metadata, yielding a per-painting correctness verdict for the title and for the artist.

Our results indicated that the model named the \emph{artist} correctly far more often than the \emph{title} (roughly $82\%$ vs.\ $28\%$): it more often identified the artist, frequently on the basis of distinctive stylistic cues, while producing a plausible but incorrect title. Figure~\ref{fig:exp2examples} shows the typical outcomes: two canonical works for which both the title and artist names are correct; one where the artist name is correct but the title corresponds to a different work by the same painter, and one where both title and artist name are incorrect.

We additionally tested whether any TAM statistic of the predicting span
anticipated its correctness: where the map concentrated (entropy), how strong it
was (mass), and how much the prediction leaned on the image rather than on the
already-generated text (its \emph{visual reliance}, the share of evidence drawn
from the image vs.\ the co-text). Map shape carried no signal---neither entropy
nor mass separated correct from wrong predictions within either type. The only
statistic with any association was visual reliance, and only for the artist:
correct artist predictions drew on average $0.93$ of their evidence from the
image against $0.85$ for wrong ones (Mann--Whitney $p \approx 0.002$, Cliff's
$\delta \approx 0.30$), i.e., wrong predictions were markedly more text-driven,
whereas for titles there was no such effect ($p \approx 0.12$). The association
was genuine but weak, and largely a painting-level property: it washed out once
visual reliance was measured relative to each artwork's own baseline, indicating
it reflected whether the description as a whole stayed visually grounded rather
than anything specific to the predicting token. As a correctness classifier it
is therefore only a triage signal (ROC-AUC $\approx 0.65$), not a reliable
predictor. We read it as a hint about \emph{why} the model is sometimes
right---it was looking at the canvas rather than confabulating from a language
prior---and as evidence that titles tend to be invented from priors while
artists can be genuinely recognized from the image.

\begin{figure}[t]
  \centering
  \begin{subfigure}{0.49\linewidth}
    \centering
    \includegraphics[width=\linewidth]{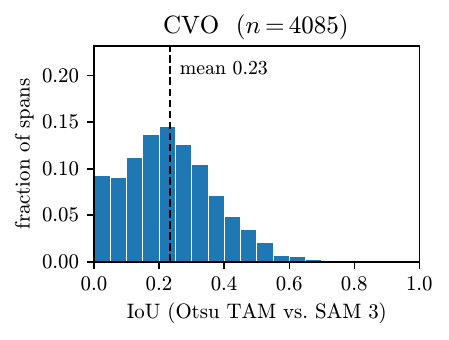}
    \caption{}
  \end{subfigure}
  \hfill
  \begin{subfigure}{0.49\linewidth}
    \centering
    \includegraphics[width=\linewidth]{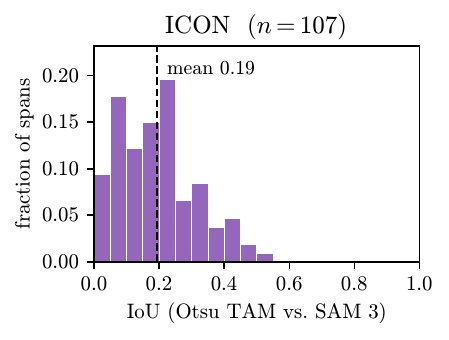}
    \caption{}
  \end{subfigure}
  \caption{IoU between the Otsu-thresholded TAM map and the SAM~3 concept mask, per category, over spans SAM~3 localized (relative frequency; dashed line = mean). Agreement is modest and higher for (a)~concrete objects than for (b)~named iconographic subjects.}
  \label{fig:exp3}
\end{figure}

\begin{figure*}[t]
  \centering
  \includegraphics[width=\linewidth]{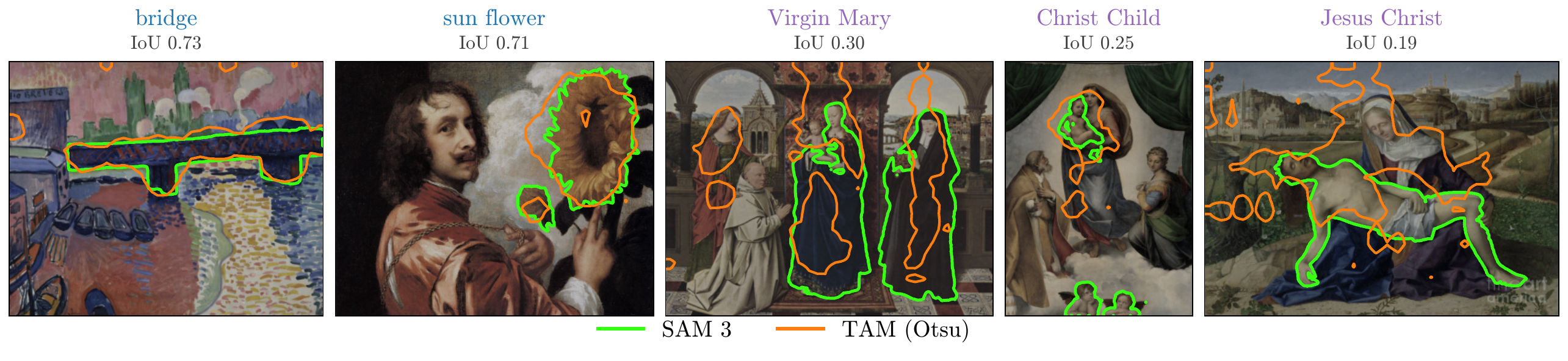}
  \caption{What each model localizes for a span (caption text above each panel, blue = \textsc{cvo}, purple = \textsc{icon}; IoU below). The painting is dimmed; the \textcolor[HTML]{2E9E0F}{green} outline is the SAM~3 concept mask, and the \textcolor[HTML]{E07000}{orange} outline is the Otsu-thresholded TAM prediction, which is coarser for both categories. The \textsc{icon} panels show characteristic errors: over-detection of \emph{Virgin Mary}; TAM correctly (although coarsely) focuses on the \emph{Christ Child} while SAM~3 also segments the cherubs; and SAM~3 cleanly segments \emph{Jesus Christ} while TAM only points at the scene center.}
  \label{fig:exp3strip}
\end{figure*}

% needs \usepackage{graphicx} and \usepackage{subcaption}
\subsection{TAM as an Open-Vocabulary Detector}
\label{sec:exp3}

The previous experiments treated the activation maps as post hoc explanations. Here we ask a more practical question: can a token activation map be repurposed as an \emph{open-vocabulary detector} for the visual arts? If a single forward pass of the captioning model already yields, for any noun phrase it produces, a spatial map that lands on the right region, then localization comes ``for free'' alongside the description, without a dedicated segmentation model.

We probed this by comparing it with SAM~3 \cite{carion2026sam}, a state-of-the-art promptable concept segmenter used as an open-vocabulary reference. For every \textsc{cvo} (concrete visual object) and \textsc{icon} (named iconographic subject) span, we prompted SAM~3 with the span's surface text and took the union of its instance masks as the reference region; we binarized the corresponding TAM map with Otsu's threshold and reported the IoU between the two. SAM~3 returned no detection for 35\% of \textsc{cvo} and 46\% of \textsc{icon} spans; lacking any reference, these were excluded from the IoU statistics.

Figure~\ref{fig:exp3} reports the IoU distributions. For both categories, the agreement is modest, and it is higher for concrete objects than for named iconographic subjects---\textsc{cvo} mean $0.233$ (median $0.224$) versus \textsc{icon} mean $0.194$ (median $0.184$). What is clear from inspecting the maps is that the Otsu-thresholded TAM predictions are consistently \emph{coarser and less precise} than the SAM~3 masks, for \textsc{cvo} as well as \textsc{icon} spans: where the two overlap, it is typically a loose TAM blob loosely enclosing a much tighter SAM region. We do not find that either method is systematically better at localizing.

The qualitative examples in Figure~\ref{fig:exp3strip} make the iconographic error modes concrete. For \emph{Virgin Mary} (third panel), both models over-detect, marking several of the depicted figures as the Virgin rather than one. For \emph{Christ Child} (fourth panel, Raphael's \emph{Sistine Madonna}) TAM---despite its coarse mask---correctly concentrates on the infant the model was attending to, whereas SAM~3, lacking the semantic context, additionally segments the two cherubs at the foot of the composition. For \emph{Jesus Christ} (fifth panel, a \emph{Piet\`a}), the situation reverses: SAM~3 cleanly segments the figure of Christ, while TAM only indicates that the model is attending to the center of the scene. Since neither approach is uniformly preferable, a promising direction for future work is to combine the semantic grounding of TAM with the precise boundaries of a dedicated segmenter to obtain better localizations than either alone.

\section{Conclusion}\label{sec:conclusions}

In this work, we studied how MLLMs visually ground different semantic components of artwork descriptions using TAM as our interpretability lens.
% Across nearly 13,000 spans from 1000 paintings, we find a clear and statistically robust pattern: concrete content --- common visual objects and iconographic figures --- is grounded in localized image regions, while abstract content --- stylistic attributes and affective expressions --- activates diffusely across the canvas. Metadata spans occupy an intermediate position, and the model proves substantially more reliable at artist attribution than title prediction, where hallucinations are frequent and difficult to anticipate from activation statistics alone. When repurposed as an open-vocabulary detector, TAM yields coarse but semantically plausible localizations with modest agreement with SAM 3 masks. Neither method dominates the other, suggesting that their complementary strengths --- semantic grounding from TAM and precise boundary delineation from dedicated segmenters --- could be combined productively to obtain localizations superior to either alone.

Several directions remain open. Extending the analysis to larger and architecturally diverse MLLMs would clarify how robust the grounding patterns are across the design space. The hallucination behavior observed in the title prediction calls for targeted mitigation strategies, such as retrieval-augmented generation or uncertainty-aware decoding, whose effectiveness could be directly measured within the TAM framework. Fusing TAM with precise segmentation models offers a promising route toward fine-grained, annotation-free iconographic localization---a practically valuable capability for digital humanities and cultural heritage applications. Finally, scaling the dataset to broader artistic traditions and underrepresented periods would test whether the grounding patterns identified here generalize beyond the Western canon.

\paragraph{Acknowledgements.} Nicola Fanelli's research is funded by a Ph.D. fellowship under the Italian ``D.M. n. 118/23'' (NRRP, Mission 4, Investment 4.1, CUP H91I23000690007).

% ============================================================
%  Bibliography
% ============================================================

\bibliographystyle{splncs04}
\bibliography{main}

\end{document}